\documentclass{article}
\usepackage{ijcai21-multiauthor}

\usepackage{times}

\usepackage{soul}
\usepackage{url}
\usepackage[hidelinks]{hyperref}
\usepackage[utf8]{inputenc}
\usepackage[small]{caption}
\usepackage{graphicx}
\usepackage{amsmath}
\usepackage{booktabs}
\usepackage{algorithm}
\usepackage{algorithmic}
\usepackage{subfigure}
\usepackage{amsthm}

\newtheorem{theorem}{Theorem}

\title{H-FL: A Hierarchical Communication-Efficient and Privacy-Protected Architecture for Federated Learning}

\author{
He Yang\footnote{Contact Author}\\
\affiliations
Xi'an Jiaotong University\\
\emails
sleepingcat@stu.xjtu.edu.cn
}

\begin{document}

\maketitle

\begin{abstract}
The longstanding goals of federated learning (FL) require rigorous privacy guarantees and low communication overhead while holding a relatively high model accuracy. However, simultaneously achieving all the goals is extremely challenging. In this paper, we propose a novel framework called \textbf{h}ierarchical \textbf{f}ederated \textbf{l}earning (H-FL) to tackle this challenge. Considering the degradation of the model performance due to the statistic heterogeneity of the training data, we devise a runtime distribution reconstruction strategy, which reallocates the clients appropriately and utilizes mediators to rearrange the local training of the clients. In addition, we design a compression-correction mechanism incorporated into H-FL to reduce the communication overhead while not sacrificing the model performance. To further provide privacy guarantees, we introduce differential privacy while performing local training, which injects moderate amount of noise into only part of the complete model. Experimental results show that our H-FL framework achieves the state-of-art performance on different datasets for the real-world image recognition tasks.
\end{abstract}

\section{Introduction}

Federated learning (FL) is a promising distributed paradigm for training a shared model while keeping all the training data localized~\cite{yang2019federated,kairouz2019advances,konevcny2016federated}. However, FL always involves expensive communication and privacy concerns in order to maintain a great model performance~\cite{li2020federated,zhang2021secure}. Therefore, how to find a great balance among model performance, communication overhead and privacy requirements is a long-term, challenging goal.

From a methodological standpoint, DGC~\cite{lin2017deep} and FetchSGD~\cite{rothchild2020fetchsgd} have given a good trade-off between communication overhead and model performance by compressing the gradients and giving some corrections. NbAFL~\cite{wei2020federated} and DP-FedAVG ~\cite{mcmahan2017learning} provide strong privacy guarantees via differential privacy without undue sacrifice on model performance. SplitNN~\cite{vepakomma2018split} can achieve higher model performance in contrast to the aforementioned methods while protecting sensitive raw data. All these works try to make some trade-offs from different perspectives. However, when treating model performance, communication overhead and privacy requirements as a whole perspective, it will introduce a completely new contradiction: the contradiction between communication overhead and privacy requirements while maintaining model performance in a certain level. Since when utilizing some privacy protection methods such as differential privacy and secure multiparty computing to provide privacy guarantees, it will inevitably introduce additional communication overhead directly or slow down the convergence rate, leading to requiring extra communication rounds for FL algorithms to converge. Therefore, we cannot just do simple combinations from different perspectives.

\begin{figure*}
\centering
\includegraphics[scale=0.78]{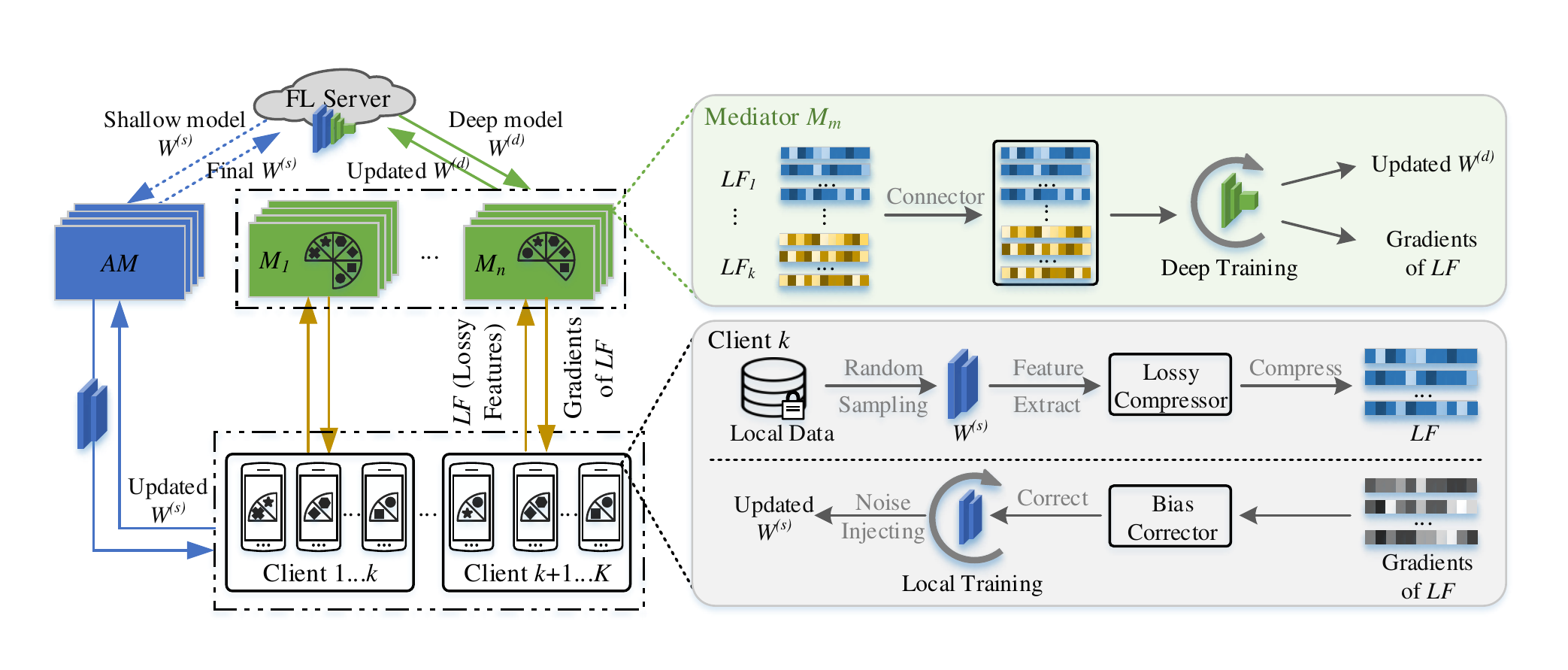}
\caption{H-FL architecture. FL server splits the complete model into two pieces: shallow model and deep model, and then distributes the former one to the \textbf{A}ggregation \textbf{M}ediator ($\mathcal{AM}$) and the latter one to the other Mediators ($\mathcal{M}$). $\mathcal{AM}$ distributes the shallow model to all the clients and is responsible for aggregating shallow models. In addition, FL server is responsible for aggregating deep models. Particularly, $\mathcal{AM}$ sends the final global shallow model to FL server at the end of the collaborative training.}
\label{H-FL}
\end{figure*}

In this paper, we develop a \textbf{h}ierarchical \textbf{f}ederated \textbf{l}earning architecture (H-FL) as shown in Figure~\ref{H-FL}. To counter-weigh the degradation of model performance due to statistic heterogeneity of the training data, H-FL introduces mediators to reconstruct the local distributions. We cluster the clients according to the KL divergence between local distributions of each client and a uniform distribution, as well as the information entropy of the local distributions, and then reallocate them to different mediators. When participating in federated tasks, H-FL selects mediators rather than clients and each mediator rearranges its clients to perform the training tasks in order to alleviate the statistic heterogeneity. In addition, we design a compression-correction mechanism to reduce the communication overhead without sacrificing the model performance, which significantly compresses the extracted features of the clients uploaded to mediators and corrects the corresponding gradients download from the mediators. To further provide privacy guarantees for clients, we introduce differential privacy when each client conducts its local training.

Our contributions can be summarized as following:
\begin{itemize}
\item To the best of our knowledge, H-FL is the first attempt to treat model performance, communication overhead and privacy requirements as a whole perspective to find a great balance among them.
\item We devise a runtime distribution reconstruction strategy to alleviate the statistic heterogeneity of the training data while not compromising user privacy. Moreover, we design a compression-correction mechanism to reduce the communication overhead without sacrificing the model performance.
\item Extensive experiments on different datasets show that our H-FL architecture achieves state-of-the-art performance on federated image recognition tasks.
\end{itemize}

\section{Related Research}

Federated learning is a collaborative distributed learning paradigm which removes the necessity to pool the raw data out from local clients. Specifically, FedAVG algorithm proposed in~\cite{mcmahan2017communication} aims to reduce the communication overhead while maintaining a good performance of the model on non-IID (Independent and identically distributed) training data, which is used as our baseline in Section~\ref{section-4}. Furthermore, concurrent works such as~\cite{lin2018dgc,sattler2019robust} have focused on further reducing communication overhead in FL via gradient sparsification, and propose solutions to counter-weigh the reduction in accuracy due to the statistic heterogeneity of the training data. Concretely, DGC~\cite{lin2018dgc} employs momentum correction and local gradient clipping on top of the gradient sparsification to ensure no loss of accuracy. In addition, DGC also uses momentum factor masking and warmup training to overcome the staleness problem caused by reduced communication. STC~\cite{sattler2019robust} propose a sparse ternary compression (STC) framework to reduce the communication overhead in FL, which enables ternarization and optimal
Golomb encoding of the weight updates and also behaves robust to non-IID training data. We conduct a comprehensive analysis and comparison with the aforementioned methods in Section~\ref{section-4} to illustrate the effectiveness of our H-FL framework. in Section~\ref{section-4} to illustrate the effectiveness of our H-FL framework.

\section{Our Approach}

In this section, we propose a hierarchical FL architecture as shown in Figure~\ref{H-FL} to find a great balance among model performance, communication overhead and privacy requirements.

\begin{table}
\centering
\begin{tabular}{cl}
\toprule
Notation & Definition \\
\midrule
$W^{(d)}_t$       & global deep model at round $t$        \\
$W^{(s)}_t$       & global shallow model at round $t$        \\
$W_t^{(c)}$    & shallow model kept in client $c$ at round $t$        \\
$W_t^{(m)}$    & deep model kept in meditor $m$ at round $t$        \\
$\mathcal{U}$       & all the clients        \\
$\mathcal{P}$       & sampling probability of each client        \\
$\mathcal{S}$       & sampling probability of each example        \\
$\mathcal{C}$       & global compression ratio        \\
$\mathcal{I}$       & iterations of deep training        \\
$\mathcal{L}$       & $\ell_2$-norm of the clipped gradients        \\
$\sigma$       & noise level        \\
\bottomrule
\end{tabular}
\caption{Notations and Definitions}
\label{tab:booktabs}
\end{table}

\subsection{Adversary Model}
We first assume that all the components (FL Server, Mediators, Clients) in H-FL have following abilities: 1) they are \textbf{\textit{honest-but-curious}}, which means that they will honestly follow the designed protocol but are curious about the others' local data; 2)  they have arbitrary auxiliary information to help infer a specific client's private information during the process of collaboratively building a shared model; 3) they do not collude with each other, which means that they will not provide any additional information to clients during the training.

\subsection{Initialization}

FL server first splits the complete model into two components: shallow model and deep model, then distributes the former one to the \textbf{A}ggregation \textbf{M}ediator ($\mathcal{AM}$) and the latter one to the other Mediators ($\mathcal{M}$). $\mathcal{AM}$ distributes the shallow model to all the clients. At the same time, FL server initializes the global hyper-parameters such as learning rate $\eta$, sampling probability of each client $\mathcal{P}$, sampling probability of each example $\mathcal{S}$, global compression ratio $\mathcal{C}$ ($\mathcal{C} < 0.5$), iterations of deep training in mediators $\mathcal{I}$, $\ell_2$-norm of the clipped gradients $\mathcal{L}$ and noise level $\sigma$. Specifically, when sampling locally in practice, we randomly permute the local data and partition them into mini-batches of the appropriate sizes for efficiency.

\subsection{Runtime Distribution Reconstruction}

In FL settings, as the training data resident in the individual clients is collected by the clients themselves on the basis of their local environments, the distribution of the local datasets will considerably differ with each other. Considering this characteristic, we redefine the optimization objective function of federated learning training on non-IID datasets as follows:
\begin{equation}
\label{eq.1}
\resizebox{.91\linewidth}{!}{$
    \displaystyle
    \left.\min _{w, p^{(c)}} \mathrm{E}_{(x, y) \sim p^{(c)}}\left[\ell ( f\left(x ; w^{(c)}\right), y\right)\right]+\sum_c \mathcal{D}_{K L}\left(p \| p^{(c)}\right)
$}
\end{equation}%
where $w^{(c)}$, $p^{(c)}$, $p$, $\mathcal{D}_{K L}$ are the weights of client \textit{c}, the local distribution of client \textit{c}, the distribution of potential global training data, KL divergence between $p$ and $p^{(c)}$, respectively. When the latter term is approximate to \textit{0}, it will degrade to an optimization problem under IID. In general FL settings, $p^{(c)}s$ are a series of different fixed distributions such that the latter term is a fixed value and the optimization objective will be consistent.

Whereas we consider $p^{(c)}s$ as variable distributions rather than fixed distributions in H-FL, so we can change local distributions arbitrarily. An intuitive way is to gather the clients' local data and form a series of different new distributions, each of which is approximate to the potential global distribution $p$, enabling the latter term in Formula (\ref{eq.1}) to be \textit{0}. However, sharing local data raises serious privacy risks and causes high communication overhead. Therefore, we introduce the runtime distribution reconstruction strategy to mitigate differences among local distributions while meeting the privacy requirements. 

Specifically, a uniform distribution $p^{(r)}$ is initialized and broadcast among the clients. Each client calculates the information entropy $\mathcal{H}^{(c)}$ of its local distribution $p^{(c)}$ and KL divergence $\mathcal{D}_{K L}(p^{(r)}\|p^{(c)})$ between $p^{(r)}$ and $p^{(c)}$. Furthermore, K-means algorithm is utilized to cluster the clients according to the binary group $(\mathcal{H}^{(c)}, \mathcal{D}_{K L}(p^{(r)}\|p^{(c)}))$. Then H-FL randomly selects clients from each cluster, marks them as a group, and assigns the group to one of mediators. The allocation pattern loops until all the clients are assigned to the corresponding mediator.

When performing local training, each client utilizes the shallow model to extract features, which will be compressed by the lossy compressor (introduced in subsection~\ref{CCM}) and sent to the corresponding mediator. After that, each mediator concatenates the received features through a connector (as shown in Figure~\ref{H-FL}) to obtain synthetic features. This procedure can be considered as sampling from a virtual reconstructed distribution $p^{(m)}$ and then conducting forward propagation using the shallow model (see Algorithm~\ref{alg:algorithm1}). Intuitively, $p^{(m)}$ will be more approximate to the potential global distribution $p$ than $p^{(c)}$. The optimization objective function will be changed to the following form:
\begin{equation}
\label{eq.2}
\begin{split}
    \min _{W, p^{(m)}} & \mathrm{E}_{(x, y) \sim p^{(m)}}\left[\ell \left( f\left(x ; W^{(c)}, W^{(m)}\right), y\right)\right]
    \\
    & + \sum_m \mathcal{D}_{K L}\left(p \| p^{(m)}\right)
\end{split}
\end{equation}

Assuming that there exists enough clients, $p^{(m)}$s will infinitely approximate the potential global distribution \textit{p} and the latter term will be \textit{0}, which is translated to the optimization problem under IID. When finishing the distribution reconstruction, each mediator leverages the synthetic features to train the deep model and then sends back the gradients of the synthesized features to the clients to assist training the shallow model. In this way, H-FL alleviates the statistic heterogeneity of the training data while not compromising user privacy.

\begin{algorithm}[tb]
\caption{Runtime distribution reconstruction}
\label{alg:algorithm1}
\textbf{Input}: $\mathcal{U}$, $\mathcal{M}$\\
\textbf{Parameter}: $W^{(c)}$,$\mathcal{P}$, $\mathcal{S}$, $\mathcal{C}$\\
\textbf{Output}: $\mathcal{B}^{(m)}$
\begin{algorithmic}[1] 
\STATE Randomly initialize a distribution $p^{(r)}$
\FOR{each $c \in \mathcal{U}$ \textbf{in parallel}}
\STATE Compute $\mathcal{H}^{(c)}$, $\mathcal{D}_{K L}(p^{(r)}\|p^{(c)})$
\ENDFOR
\STATE Cluster according to $(\mathcal{H}^{(c)}, \mathcal{D}_{K L}(p^{(r)}\|p^{(c)}))$
\FOR{each $m \in \mathcal{M}$}
\STATE Randomly select clients from each cluster according to the same ratio $1 / \left|\mathcal{M}\right|$ and assign them to \textit{m}
\STATE $\mathcal{B}^{(m)} \leftarrow \emptyset $
\ENDFOR
\STATE $\mathcal{M}^{t} \leftarrow$ (Randomly sampling mediators in $\mathcal{M}$)
\FOR{each $m \in \mathcal{M}^{t}$}
\STATE $\mathcal{U}^t \leftarrow$ (Randomly sampling clients in $\mathcal{U}$  with $\mathcal{P}$)
\FOR{each $c \in \mathcal{U}^t$}
\STATE Randomly sampling a mini-batch $X^{(c)}$ with $S$
\STATE $\mathcal{O}^{(c)} \leftarrow W^{(c)}X^{(c)}$
\STATE $k \leftarrow \lfloor \left| \mathcal{O}^{(c)} \right| * C\rfloor$
\STATE $\mathcal{B}^{(m)} \leftarrow \mathcal{B}^{(m)} \cup LF(\mathcal{O}^{(c)})$
\ENDFOR
\ENDFOR
\STATE \textbf{return} $\mathcal{B}^{(m)}$
\end{algorithmic}
\end{algorithm}

\subsection{Compression-Correction Mechanism}
\label{CCM}

To reduce the communication overhead, each participating client compresses the extracted features through the lossy compressor in Figure~\ref{H-FL} by:
\begin{equation}
\label{eq.3}
    LF(\mathcal{O}) = U_{\mathcal{O}}[:, :k] \Sigma_{\mathcal{O}}[:k] V^T_{\mathcal{O}}[:k]
\end{equation}
where $\mathcal{O}$ is feature matrix extracted by the shallow model, $U_{\mathcal{O}}$, $\Sigma_{\mathcal{O}}$ and $V^T_{\mathcal{O}}$ are the results of singular value decomposition (SVD) respectively, $U_{\mathcal{O}}[:, :k]$, $\Sigma_{\mathcal{O}}[:k]$ and $V_{\mathcal{O}}[:k]$ represent the first $k$ columns of $U_{\mathcal{O}}$, $\Sigma_{\mathcal{O}}$ and $V_{\mathcal{O}}$ respectively. In this way, the feature matrix can be changed to a low-rank matrix that can be expressed by as the product of two relatively small matrices, thus reducing the communication overhead.

For the sake of clarification, let us introduce some new representations:
\begin{equation}
\label{eq.4}
    \begin{array}{l}
        \mathcal{O}=W^{(c)}X^{(c)} \\
        \mathcal{B}=LF(\mathcal{O}) \\
        \mathcal{A}=W^{(m)}\mathcal{B} \\
        \mathcal{L}=\mathrm{E}[\ell(\mathcal{A}, y)]
    \end{array}
\end{equation}
When updating $W^{(c)}$, we should compute $dW^{(c)}$ as follows according to the chain rule:
\begin{equation}
\label{eq.5}
    d W^{(c)}=\frac{\partial \mathcal{L}}{\partial \mathcal{A}} \cdot \frac{\partial \mathcal{A}}{\partial \mathcal{B}} \cdot \frac{\partial \mathcal{B}}{\partial W^{(c)}}
\end{equation}

However, according to formula (\ref{eq.3}), we cannot compute $\partial \mathcal{B} / \partial W^{(c)}$ directly since there is no direct differentiable mapping from $W^{(c)}$ to $\mathcal{B}$. For convenience, $\partial \mathcal{O} / \partial W^{(c)}$ can be used instead of $\partial \mathcal{B} / \partial W^{(c)}$, which may still work but it leads to a reduction in model accuracy.

Therefore, we design a bias corrector on clients to correct the gradients of lossy features, which could build the mapping from $\mathcal{O}$ to $\mathcal{B}$ so that we can better approximate $\partial \mathcal{B} / \partial W^{(c)}$ and counter-weigh the reduction. According to the feature of SVD, we can get:
\begin{equation}
\label{eq.6}
    \mathcal{B}=U_{\mathcal{O}} \mathcal{D}_{k} U_{\mathcal{O}}^{T}\mathcal{O}
\end{equation}
where $U_{\mathcal{O}}$ here is just the same thing as the $U_{\mathcal{O}}$ in formula (\ref{eq.3}), $\mathcal{D}_{k}$ represents a diagonal matrix where its first $k$ elements on the diagonal are \textit{1} and the rest are \textit{0}. Therefore, $\partial \mathcal{B} / \partial W^{(c)}$ can be rewritten as: 
\begin{equation}
\label{eq.7}
    \partial \mathcal{B} / \partial {W}^{(c)} \approx U_{\mathcal{O}} \mathcal{D}_{k} U_{\mathcal{O}}^{T} \cdot  (\partial \mathcal{O} / \partial W^{(c)})
\end{equation}
Thus, the bias corrector can be considered as consisting of many fully connected layers stacked on top of each other, and the parameters depend on the SVD results of the features extracted from the shallow model. In other words, the parameters of the bias corrector will be updated during the procedure of forward propagation. We also compare the results for the presence or absence of the bias corrector through appropriate experiments in Section~\ref{section-4}.

\begin{algorithm}[tb]
\caption{The workflow for H-FL}
\label{alg:algorithm2}
\textbf{Input}: $\mathcal{U}$, $\mathcal{AM}$, $\mathcal{M}$\\
\textbf{Parameter}: $W^{(m)}_t$, $W^{(c)}_t$, $\mathcal{P}$, $\mathcal{S}$, $\mathcal{C}$, $\mathcal{I}$, $\mathcal{L}$, $\sigma$\\
\textbf{Output}: $W^{(d)}_{t+1}$, $W^{(s)}_{t+1}$ \\
\textit{\textbf{Mediators}}:
\begin{algorithmic}[1] 
\STATE $\mathcal{B}^{(m)} \leftarrow$ Run-time data augmentation
\FOR{each $m \in \mathcal{M} \setminus \mathcal{AM}$ \textbf{in parallel}}
\FOR{each epoch $i$ from 1 to $\mathcal{I}$}
\STATE $W^{(m)}_t \leftarrow W^{(m)}_t - \eta\nabla_{W^{(m)}_t}\ell(W^{(m)}_t \mathcal{B}^{(m)},y)$
\ENDFOR
\STATE $d\mathcal{B}^{(m)} \leftarrow \nabla_{\mathcal{B}^{(m)}}\ell(W^{(m)}_t \mathcal{B}^{(m)},y)$
\FOR{each $c \in m$}
\STATE $d\mathcal{B}^{(c)} \leftarrow d\mathcal{B}^{(m)}[:n^{(c)}]$
\STATE $d\mathcal{B}^{(m)} \leftarrow d\mathcal{B}^{(m)}[n^{(c)}:]$
\ENDFOR
\ENDFOR
\end{algorithmic}
\textit{\textbf{Clients}}:
\begin{algorithmic}[1]
\FOR{each $c \in \mathcal{U}^{t}$ \textbf{in parallel}}
\STATE $\mathcal{B}^{(c)} \leftarrow U_{\mathcal{O}}^{(c)} D_{k}^{(c)} U_{\mathcal{O}}^{(c) T}\mathcal{O}^{(c)}$
\STATE $dW^{(c)}_t \leftarrow d\mathcal{B}^{(c)} \partial \mathcal{B}^{(c)} / \partial W^{(c)}_t$
\STATE $dW^{(c)}_t \leftarrow dW^{(c)}_t + \mathcal N\left(0, \sigma^{2} \mathcal{L}^{2} I / n^{(c)}\right)$
\STATE $W^{(c)}_t \leftarrow W^{(c)}_t - \eta dW^{(c)}_t$
\ENDFOR
\end{algorithmic}
\textit{\textbf{FL Server}}:
\begin{algorithmic}[1]
\STATE $W^{(d)}_{t+1} \leftarrow \frac{\sum_{m \in \mathcal{M} \setminus AM} W^{(m)}_t}{|\mathcal{M} \setminus AM|}$
\end{algorithmic}
\textit{\textbf{AM}}:
\begin{algorithmic}[1]
\STATE $W^{(s)}_{t+1} \leftarrow \frac{\sum_{c \in \mathcal{U}^{t}} W^{(c)}_t}{\left|\mathcal{U}^{t}\right|}$
\end{algorithmic}
\end{algorithm}

\begin{figure*}[!t]
 \subfigure[Methods On FMNIST]{
  \begin{minipage}[t]{0.32\linewidth}
   \centering
   \includegraphics[width = 1\textwidth]{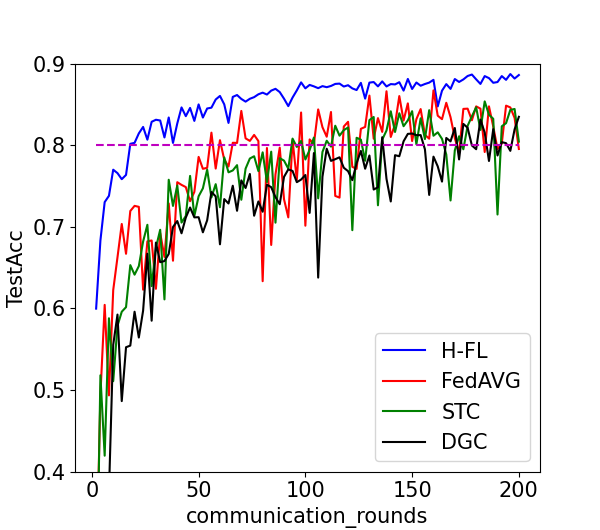}
  \end{minipage}
  \label{Methods On FMNIST}
 }
 \subfigure[Methods On CIFAR10]{
  \begin{minipage}[t]{0.32\linewidth}
   \centering
   \includegraphics[width = 1\textwidth]{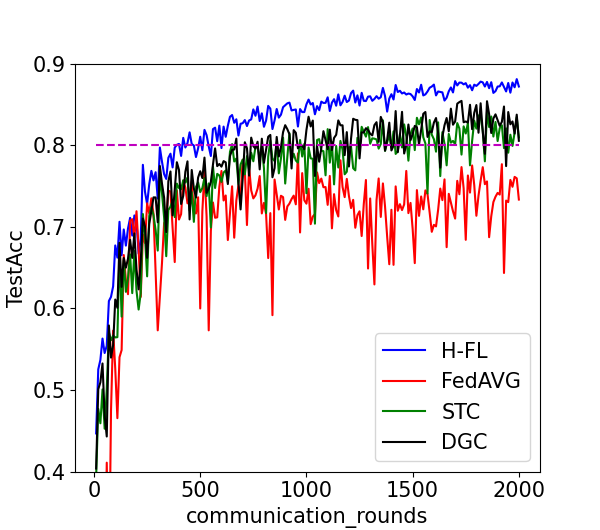}
  \end{minipage}
  \label{Methods On CIFAR10}
 }
 \subfigure[Participating Ratio $\mathcal{P}$]{
  \begin{minipage}[t]{0.32\linewidth}
   \centering
   \includegraphics[width = 1\textwidth]{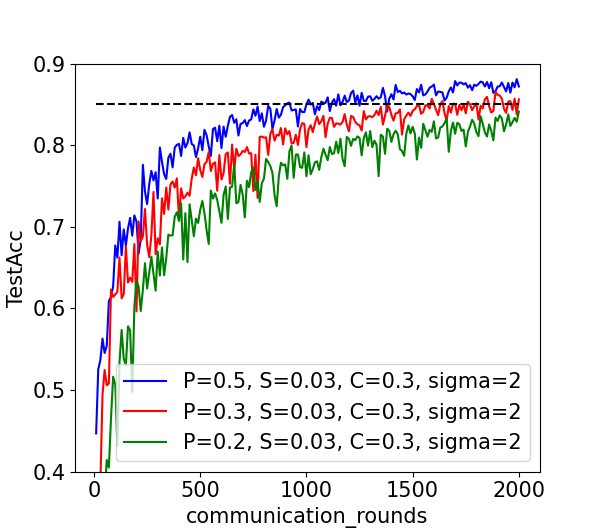}
  \end{minipage}
  \label{Participating Ratio}
 }
 \subfigure[Sampling Probability $\mathcal{S}$]{
  \begin{minipage}[t]{0.32\linewidth}
   \centering
   \includegraphics[width = 1\textwidth]{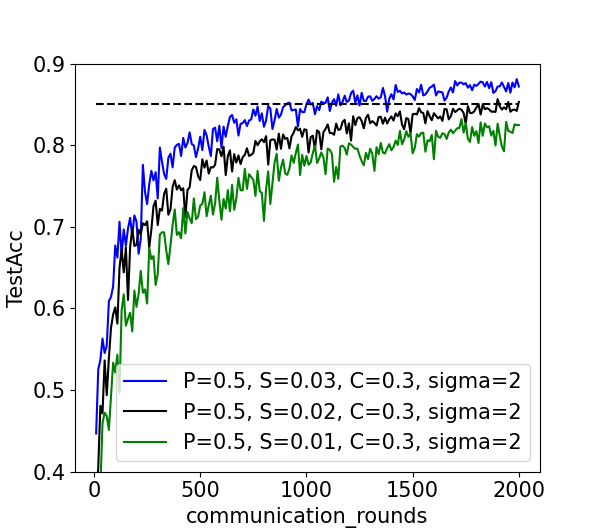}
  \end{minipage}
  \label{Sampling Probability}
 }
 \subfigure[Compression Ratio $\mathcal{C}$]{
  \begin{minipage}[t]{0.32\linewidth}
   \centering
   \includegraphics[width = 1\textwidth]{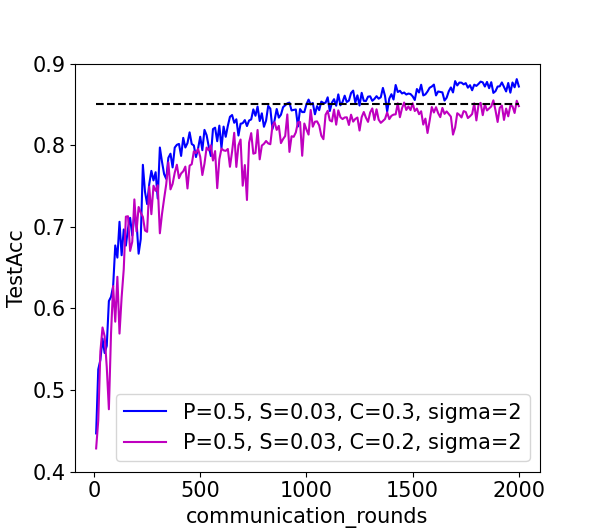}
  \end{minipage}
  \label{Compression Ratio}
 }
 \subfigure[Noise Level $\sigma$]{
  \begin{minipage}[t]{0.32\linewidth}
   \centering
   \includegraphics[width = 1\textwidth]{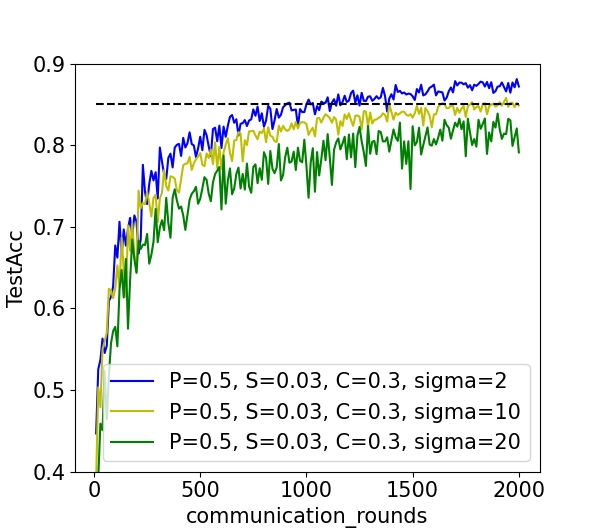}
  \end{minipage}
  \label{Noise Level}
 }
 \caption{Behavior of the Model Performance and Influence of Different Parameters for H-FL.}
 \label{Parameters}
\end{figure*}

After we obtain rectified $dW^{(c)}$, we conduct gradient clipping so that the $\ell_2$ norm of $dW^{(c)}$ is limited to $L$ and then add noise for it in order to protect privacy:
\begin{equation}
\label{eq.8}
    g^{(c)} \leftarrow \frac{g^{(c)}}{\max \left(1,\left\|g^{(c)}\right\|_{2} / L\right)}+\mathcal N\left(0, \frac{\sigma^{2} L^{2} I}{n^{(c)}}\right)
\end{equation}

where $g^{(c)}$ is $dW^{(c)}$ itself, $n^{(c)}$ is the size of the sampled mini-batch in client $c$, $\mathcal N$ is the Gaussian distribution with mean \textit{0} and standard deviation $\sigma L I / \sqrt{n}$.

In summary, the workflow of H-FL mainly includes runtime distribution reconstruction, training and aggregation, the pseudo-code of which is given as Algorithm~\ref{alg:algorithm2}.

\begin{theorem}
    Formula (\ref{eq.8}) satisfies differential privacy in distribute environment and the privacy loss can be tracked via moments accountant.
\end{theorem}

\renewcommand{\qedsymbol}{}
\begin{proof}
We can consider the first term of formula (\ref{eq.8}) as follows approximately:
\begin{equation}
\label{eq.9}
    g^{(c)} = \frac{\sum_{i=1}^{n^{(c)}} g\left(x_{i}^{(c)}\right) / \max \left(1,\left\|g\left(x_{i}^{(c)}\right)\right\|_{2} / L\right)}{n^{(c)}}
\end{equation}

where \textit{g} is the gradient of backward propagation, $x_{i}^{(c)}$ is the \textit{i}-th example of client \textit{c} and $n^{(c)}$ is the size of sampled mini-batch of client \textit{c}. In addition, we can also consider the latter term of formula (\ref{eq.8}) as follows according to central limit theorem:
\begin{equation}
\label{eq.10}
    \mathcal N\left(0, \frac{\sigma^{2} L^{2} I}{n^{(c)}}\right) = \frac{\sum_{i=1}^{n^{(c)}}\mathcal N\left(0, \sigma^{2} L^{2} I\right)}{n^{(c)}}
\end{equation}

Therefore, we can rewrite formula (\ref{eq.8}) as formula (\ref{eq.11}), which satisfies example-level differential privacy for each client according to Theorem 2~\cite{abadi2016deep}. In addition, since $L$ and $\sigma$ are the same for all clients, the privacy loss accumulated via moment accountant for each client in the distributed environment is the same. It also satisfies differential privacy in the distributed environment according to differential privacy parallel principle.
\begin{equation}
\label{eq.11}
\resizebox{.91\linewidth}{!}{$
    \displaystyle
    g^{(c)} = \frac{\sum_{i=1}^{n^{(c)}} \frac{g\left(x_{i}^{(c)}\right)}{\max \left(1,\left\|g\left(x_{i}^{(c)}\right)\right\|_{2} / L\right)}  + \mathcal N\left(0, \sigma^{2} L^{2} I\right)}{n^{(c)}}
$}
\end{equation}%

\end{proof}

\section{Experimental Results}
\label{section-4}

We evaluate H-FL on different datasets and compare the performance to FedAVG~\cite{mcmahan2017communication}, STC~\cite{sattler2019robust} and DGC~\cite{lin2018dgc} in non-IID environments. Specifically, we have trained a modified version of \textit{LeNet5}~\cite{lecun1998gradient} network on FMNIST~\cite{xiao2017fashion} and a modified \textit{VGG16}~\cite{simonyan2014very} network network on \textit{CIFAR10}~\cite{krizhevsky2009learning} respectively. In addition, the first two CNN blocks of VGG16 and the first one CNN block of modified LeNet5 are set to the shallow part in practice. All the batch-normalization layers are removed in the shallow model. The experiment settings are listed in Table~\ref{experiment settings}.

\begin{table}[!htbp]
\centering
\resizebox{.91\linewidth}{!}{
\begin{tabular}{ccccccc}
\toprule
Dataset & Clients & Mediators & $\eta$ & classes per client & $\mathcal{I}$ & $\mathcal{L}$\\
\midrule
CIFAR10 & 100 & 3 & 0.015 & 3 & 10 & 1\\
\midrule
FMNIST & 100 & 3 & 0.015 & 2 & 10 & 1\\
\bottomrule
\end{tabular}
}
\caption{Experiment Settings}
\label{experiment settings}
\end{table}

\subsection{Behavior Of The Model Performance}

\begin{figure*}[!t]
 \subfigure[Effectiveness of Bias Corrector]{
  \begin{minipage}[t]{0.32\linewidth}
   \centering
   \includegraphics[width = 1\textwidth]{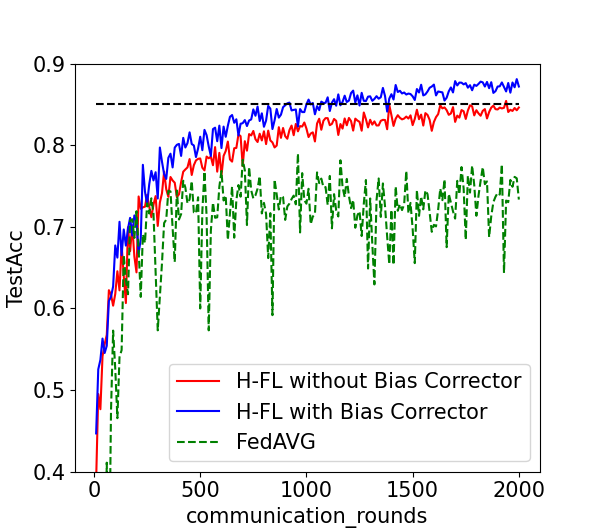}
  \end{minipage}
  \label{Bias Corrector}
 }
 \subfigure[Communication Overhead On FMNIST]{
  \begin{minipage}[t]{0.32\linewidth}
   \centering
   \includegraphics[width = 1\textwidth]{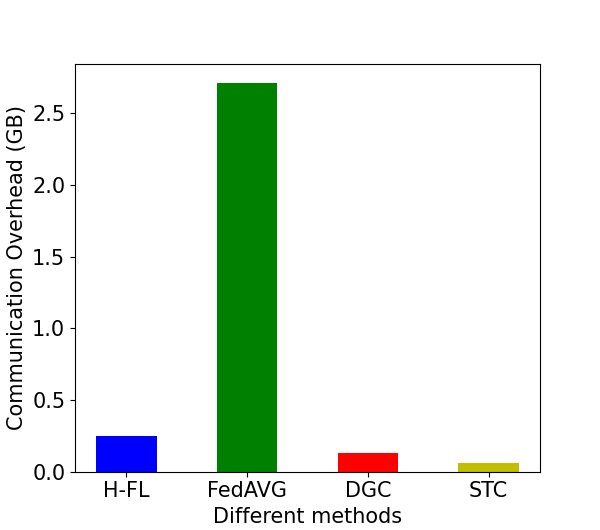}
  \end{minipage}
  \label{FMNST Overhead}
  }
  \subfigure[Communication Overhead On CIFAR10]{
  \begin{minipage}[t]{0.32\linewidth}
   \centering
   \includegraphics[width = 1\textwidth]{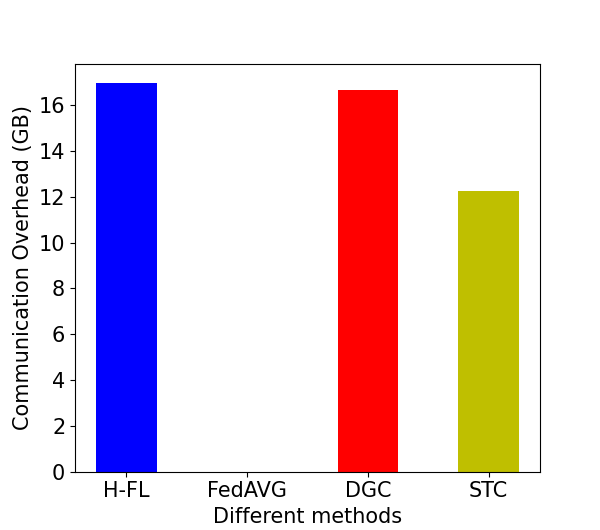}
  \end{minipage}
  \label{CIFAR10 Overhead}
 }
 \caption{Effectiveness of the Bias Corrector and Communication Overhead}
\end{figure*}
Figure~\ref{Methods On FMNIST} and Figure~\ref{Methods On CIFAR10} show the top-1 accuracy of \textit{LeNet-5} on FMNIST after 200 communication rounds and the accuracy of \textit{VGG16} on \textit{CIFAR10} after 2000 communication rounds respectively using  H-FL and the aforementioned three methods. The magenta dotted line refers to an accuracy of 80\%. The experiment results show that H-FL outperforms the other methods both on convergence rate and final accuracy. The results are quite reasonable since H-FL reconstructs a series of virtual distributions $p^{(m)}$s, each of which is more closer to the potential global distribution and the optimization problem under non-IID is almost turn into that under IID. Thus, H-FL have the better capability to handle the heterogeneous dataset. Specifically, we take the average of the last 10 rounds of the accuracy as the final accuracy after 200 rounds on FMNIST and 2000 rounds on CIFAR10 respectively. H-FL achieves an accuracy of 88.16\% on FMNIST, whereas FedAVG, DGC and STC only achieve 82.28\%, 82.00\%, and 82.12\% respectively. Moreover H-FL achieves an accuracy of 87.28\% on CIFAR10, whereas FedAVG, DGC and STC only achieve 73.83\%, 81.25\% and 81.24\%.

\subsection{Influence Of Different Parameters For H-FL}
From Figure~\ref{Participating Ratio}, Figure~\ref{Sampling Probability} and Figure~\ref{Compression Ratio}, we can observe that as $\mathcal{P}$, $\mathcal{S}$ and $\mathcal{C}$ increase, the model performance and the convergence behavior are getting better. The phenomenon is quite reasonable because: 1) As aforementioned, the procedure of reconstructing distributions in H-FL is closely related to the training samples of each client. The larger $\mathcal{P}$ and $\mathcal{S}$ are, the more the training samples are, and the closer the reconstructed distribution is to the potential global distribution, thus reducing the impact of non-IID and obtaining a relatively ideal effect; 2) The number of training samples of each client has a great impact on noise injecting. As $\mathcal{S}$ increase, the number of training samples of each client becomes larger, so that the $n^{(c)}$ in Formula~\ref{eq.8} will be larger and the injected noise is correspondingly small; 3) As $\mathcal{C}$ increases, the lossy compression becomes less and less effective and the behavior will get better. In addition, as the noise level $\sigma$ increases in Figure~\ref{Noise Level}, the oscillation amplitude of the accuracy curve becomes larger, the convergence speed becomes slower and the model performance becomes worse.

\subsection{Effectiveness Of The Bias Corrector}

Figure~\ref{Bias Corrector} shows the top-1 accuracy of \textit{VGG16} on \textit{CIFAR10} with and without the bias corrector. The black dotted line refers to an accuracy of 85\%. As we can see, bias corrector has significant influence on the convergence behavior and the final accuracy of H-FL. When there exists the bias corrector, the accuracy of the global model converges to 85\% around 1000 rounds. Whereas when we remove the bias corrector, the accuracy gradually approximates to 85\% until 2000 rounds. Additionally, we take the average of the last 10 rounds of the accuracy as the final accuracy after 2000 rounds, and the bias corrector can achieve an improvement of 2.47 percentage points. The result of the experiment is in line with our expectation since the bias corrector gives a relatively precise approximation of $dW^{(c)}$ when it can't be calculated directly, and once we remove the bias corrector, it will obtain a biased $dW^{(c)}$, leading to the degradation in model performance and other metrics (convergence behavior).

\subsection{Communication Overhead}

Finally, we compare the different methods with respect to the communication overhead which are required to achieve a certain target accuracy. As we can see in the Figure~\ref{Methods On FMNIST} and Figure~\ref{Methods On CIFAR10}, the convergence behavior is much better than other methods, which considerably reduces the communication rounds. Notice that FedAVG does not converge on \textit{CIFAR10}, thus we do not show that in Figure~\ref{CIFAR10 Overhead}. To compare the communication overhead, we set a window of size 10, which is utilized to calculate an average of 10 rounds. The communication overhead accumulates while moving forward the window until the average accuracy is no less than the target accuracy (80\% in our experiments). Figure~\ref{FMNST Overhead} and Figure~\ref{CIFAR10 Overhead} show the communication overhead required to achieve the target accuracy for the different methods on \textit{FMNIST} and \textit{CIFAR10} respectively.

\section{Conclusion}

In this paper, we present a \textbf{H}ierarchical \textbf{F}ederated \textbf{L}earning architecture (H-FL) to find a great balance among model performance, communication overhead and privacy requirements. Firstly, we devise a runtime distribution reconstruction strategy to counter-weigh the degradation due to non-IID. Then we design a compression-correction mechanism to reduce the communication overhead without sacrificing the model performance. The experimental results have proved that H-FL achieves the state-of-the-art performance under different federated learning tasks.

\section*{Acknowledgements}

We would like to thank anonymous reviewers for their helpful comments. This work was supported by the the National Key Research and Development Program of China (No. 2018AAA0100500),  NSFC Grant No. 61872285, 62072367, 61772413, the Key Research and Development Program of Shaanxi Province (2020GY-033), Key Science and Technology Project of Henan Province (201300210400), Fundamental Research Funds for the Central Universities (xzy012020112).

\bibliographystyle{ijcai21-multiauthor}
\bibliography{ijcai21-multiauthor}

\begin{thebibliography}{}

\bibitem[\protect\citeauthoryear{Abadi \bgroup \em et al.\egroup
  }{2016}]{abadi2016deep}
Martin Abadi, Andy Chu, Ian Goodfellow, H~Brendan McMahan, Ilya Mironov, Kunal
  Talwar, and Li~Zhang.
\newblock Deep learning with differential privacy.
\newblock In {\em Proceedings of the 2016 ACM SIGSAC Conference on Computer and
  Communications Security}, pages 308--318, 2016.

\bibitem[\protect\citeauthoryear{Kairouz \bgroup \em et al.\egroup
  }{2019}]{kairouz2019advances}
Peter Kairouz, H~Brendan McMahan, Brendan Avent, Aur{\'e}lien Bellet, Mehdi
  Bennis, Arjun~Nitin Bhagoji, Keith Bonawitz, Zachary Charles, Graham Cormode,
  Rachel Cummings, et~al.
\newblock Advances and open problems in federated learning.
\newblock {\em arXiv preprint arXiv:1912.04977}, 2019.

\bibitem[\protect\citeauthoryear{Kone{\v{c}}n{\`y} \bgroup \em et al.\egroup
  }{2016}]{konevcny2016federated}
Jakub Kone{\v{c}}n{\`y}, H~Brendan McMahan, Felix~X Yu, Peter Richt{\'a}rik,
  Ananda~Theertha Suresh, and Dave Bacon.
\newblock Federated learning: Strategies for improving communication
  efficiency.
\newblock {\em arXiv preprint arXiv:1610.05492}, 2016.

\bibitem[\protect\citeauthoryear{Krizhevsky \bgroup \em et al.\egroup
  }{2009}]{krizhevsky2009learning}
Alex Krizhevsky, Geoffrey Hinton, et~al.
\newblock Learning multiple layers of features from tiny images.
\newblock {\em Master's thesis, University of Tront}, 2009.

\bibitem[\protect\citeauthoryear{LeCun \bgroup \em et al.\egroup
  }{1998}]{lecun1998gradient}
Yann LeCun, L{\'e}on Bottou, Yoshua Bengio, and Patrick Haffner.
\newblock Gradient-based learning applied to document recognition.
\newblock {\em Proceedings of the IEEE}, 86(11):2278--2324, 1998.

\bibitem[\protect\citeauthoryear{Li \bgroup \em et al.\egroup
  }{2020}]{li2020federated}
Tian Li, Anit~Kumar Sahu, Ameet Talwalkar, and Virginia Smith.
\newblock Federated learning: Challenges, methods, and future directions.
\newblock {\em IEEE Signal Processing Magazine}, 37(3):50--60, 2020.

\bibitem[\protect\citeauthoryear{Lin \bgroup \em et al.\egroup
  }{2017}]{lin2017deep}
Yujun Lin, Song Han, Huizi Mao, Yu~Wang, and William~J Dally.
\newblock Deep gradient compression: Reducing the communication bandwidth for
  distributed training.
\newblock {\em arXiv preprint arXiv:1712.01887}, 2017.

\bibitem[\protect\citeauthoryear{Lin \bgroup \em et al.\egroup
  }{2018}]{lin2018dgc}
Yujun Lin, Song Han, Huizi Mao, Yu~Wang, and William~J Dally.
\newblock {Deep Gradient Compression: Reducing the communication bandwidth for
  distributed training}.
\newblock In {\em The International Conference on Learning Representations},
  2018.

\bibitem[\protect\citeauthoryear{McMahan \bgroup \em et al.\egroup
  }{2017a}]{mcmahan2017communication}
Brendan McMahan, Eider Moore, Daniel Ramage, Seth Hampson, and Blaise~Aguera
  y~Arcas.
\newblock Communication-efficient learning of deep networks from decentralized
  data.
\newblock In {\em Artificial Intelligence and Statistics}, pages 1273--1282.
  PMLR, 2017.

\bibitem[\protect\citeauthoryear{McMahan \bgroup \em et al.\egroup
  }{2017b}]{mcmahan2017learning}
H~Brendan McMahan, Daniel Ramage, Kunal Talwar, and Li~Zhang.
\newblock Learning differentially private recurrent language models.
\newblock {\em arXiv preprint arXiv:1710.06963}, 2017.

\bibitem[\protect\citeauthoryear{Rothchild \bgroup \em et al.\egroup
  }{2020}]{rothchild2020fetchsgd}
Daniel Rothchild, Ashwinee Panda, Enayat Ullah, Nikita Ivkin, Ion Stoica,
  Vladimir Braverman, Joseph Gonzalez, and Raman Arora.
\newblock Fetchsgd: Communication-efficient federated learning with sketching.
\newblock In {\em International Conference on Machine Learning}, pages
  8253--8265. PMLR, 2020.

\bibitem[\protect\citeauthoryear{Sattler \bgroup \em et al.\egroup
  }{2019}]{sattler2019robust}
Felix Sattler, Simon Wiedemann, Klaus-Robert M{\"u}ller, and Wojciech Samek.
\newblock Robust and communication-efficient federated learning from non-iid
  data.
\newblock {\em IEEE transactions on neural networks and learning systems},
  2019.

\bibitem[\protect\citeauthoryear{Simonyan and
  Zisserman}{2014}]{simonyan2014very}
Karen Simonyan and Andrew Zisserman.
\newblock Very deep convolutional networks for large-scale image recognition.
\newblock {\em arXiv preprint arXiv:1409.1556}, 2014.

\bibitem[\protect\citeauthoryear{Vepakomma \bgroup \em et al.\egroup
  }{2018}]{vepakomma2018split}
Praneeth Vepakomma, Otkrist Gupta, Tristan Swedish, and Ramesh Raskar.
\newblock Split learning for health: Distributed deep learning without sharing
  raw patient data.
\newblock {\em arXiv preprint arXiv:1812.00564}, 2018.

\bibitem[\protect\citeauthoryear{Wei \bgroup \em et al.\egroup
  }{2020}]{wei2020federated}
Kang Wei, Jun Li, Ming Ding, Chuan Ma, Howard~H Yang, Farhad Farokhi, Shi Jin,
  Tony~QS Quek, and H~Vincent Poor.
\newblock Federated learning with differential privacy: Algorithms and
  performance analysis.
\newblock {\em IEEE Transactions on Information Forensics and Security}, 2020.

\bibitem[\protect\citeauthoryear{Xiao \bgroup \em et al.\egroup
  }{2017}]{xiao2017fashion}
Han Xiao, Kashif Rasul, and Roland Vollgraf.
\newblock Fashion-mnist: a novel image dataset for benchmarking machine
  learning algorithms.
\newblock {\em arXiv preprint arXiv:1708.07747}, 2017.

\bibitem[\protect\citeauthoryear{Yang \bgroup \em et al.\egroup
  }{2019}]{yang2019federated}
Qiang Yang, Yang Liu, Tianjian Chen, and Yongxin Tong.
\newblock Federated machine learning: Concept and applications.
\newblock {\em ACM Transactions on Intelligent Systems and Technology (TIST)},
  10(2):1--19, 2019.

\bibitem[\protect\citeauthoryear{Zhang \bgroup \em et al.\egroup
  }{2021}]{zhang2021secure}
Qingsong Zhang, Bin Gu, Cheng Deng, and Heng Huang.
\newblock Secure bilevel asynchronous vertical federated learning with backward
  updating.
\newblock {\em arXiv preprint arXiv:2103.00958}, 2021.

\end{thebibliography}

\end{document}